\documentclass[10pt,twocolumn,letterpaper]{article}

\usepackage{iccv}
\usepackage{times}
\usepackage{epsfig}
\usepackage{graphicx}
\usepackage{amsmath}
\usepackage{amssymb}
\usepackage{subcaption}
\usepackage{multirow}
\usepackage{color}
\usepackage{enumitem}
\setlist{nolistsep}

\usepackage[pagebackref=true,breaklinks=true,letterpaper=true,colorlinks,bookmarks=false]{hyperref}

\iccvfinalcopy 

\def\iccvPaperID{1139} 

\ificcvfinal\fi
\begin{document}

\title{See, Hear, and Read: Deep Aligned Representations} 

\author{Yusuf Aytar, Carl Vondrick, Antonio Torralba\\
Massachusetts Institute of Technology\\
\texttt{\{yusuf,vondrick,torralba\}@csail.mit.edu}}

\maketitle

\begin{abstract}

We capitalize on large amounts of readily-available, synchronous data to learn a deep discriminative representations shared across three major natural modalities: vision, sound and language.
By leveraging over a year of sound from video and millions of sentences paired with images, we jointly train a deep convolutional network for aligned representation learning. 
Our experiments suggest that this representation is useful for several tasks, such as cross-modal retrieval or transferring classifiers between modalities. Moreover, although our network is only trained with image+text and image+sound pairs, it can transfer between text and sound as well, a transfer the network never observed during training. 
Visualizations of our representation reveal many hidden units which automatically emerge to detect concepts, independent of the modality. 

\end{abstract}

\section{Introduction}

Invariant representations are core for vision, audio, and language models because they abstract our data. For example, we desire viewpoint and scale invariance in vision, reverberation and background noise invariance in audio, and synonym and grammar invariance in language. Discriminative, invariant representations learned from large datasets have enabled machines to understand unconstrained situations to huge success \cite{krizhevsky2012imagenet,mikolov2013distributed,hannun2014deep,aytar2016soundnet}. 

The goal of this paper is to create representations that are robust in another way: we learn representations that are aligned across modality. Consider the sentence ``she jumped into the pool.'' This same concept could also appear visually or aurally, such as the image of a pool or the sound of splashing. Representations are robust to modality if the there is alignment in the representation across  modalities. The pool image, the splashing sound, and the above sentence should have similar representations.

\begin{figure}[t]
    \centering
    \includegraphics[width=\linewidth]{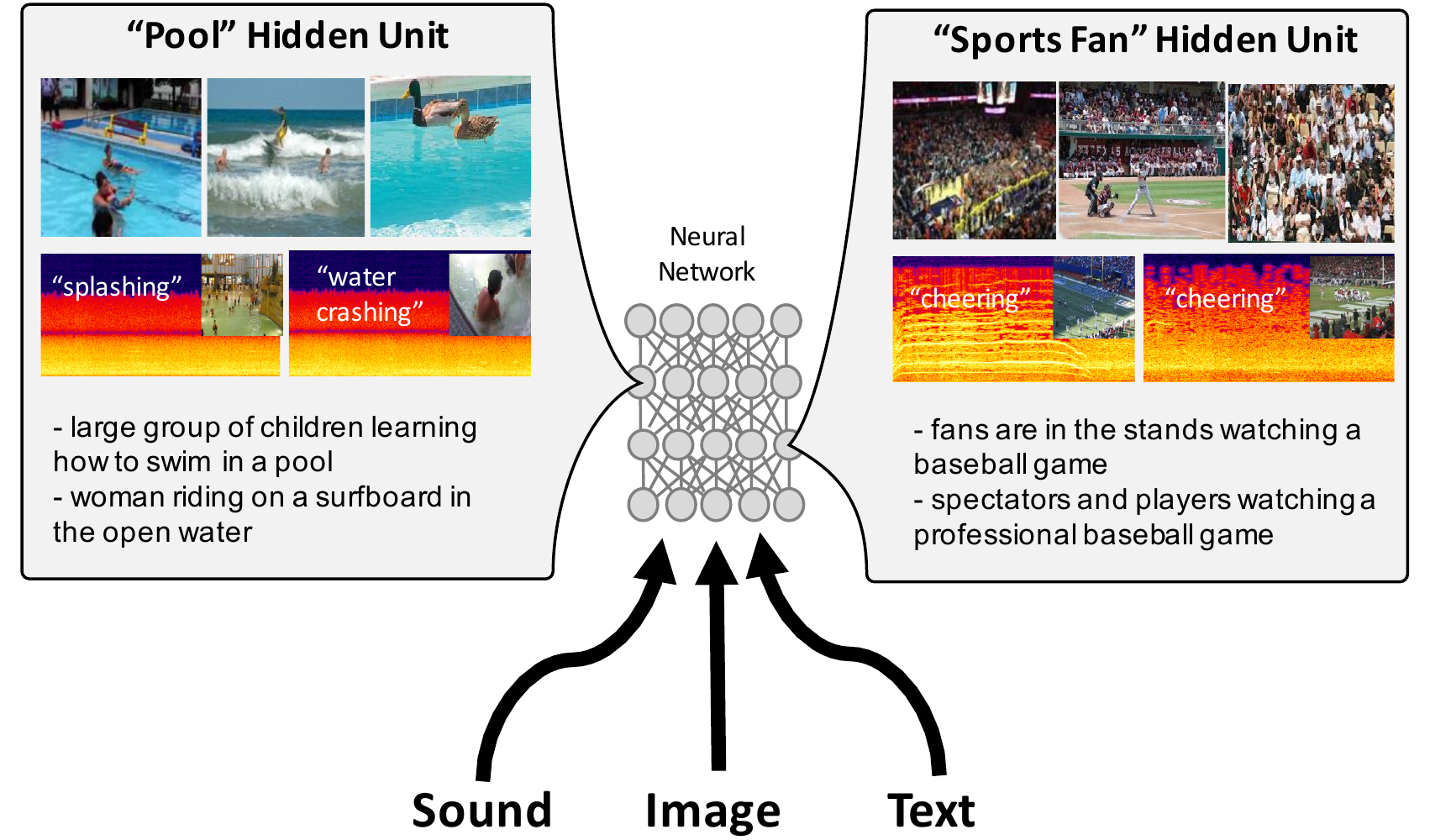}
    \caption{\textbf{Aligned Representations:} We present a deep cross-modal convolutional network that learns a representation that is aligned across three senses: seeing, hearing, and reading. Above, we show  inputs that activate a hidden unit the most. Notice that units fire on concepts independent of the modality. See Figure \ref{fig:neuron} for more.}
    \label{fig:teaser}
\end{figure}

We believe aligned cross-modal representations will have a large impact in computer vision because they are fundamental components for machine perception to understand relationships between modalities. Cross-modal perception plays key roles in the human perceptual system to recognize concepts in different modalities \cite{calvert2000evidence,giard1999auditory}. Cross-modal representations also have many practical applications in recognition and graphics, such as transferring learned knowledge between modalities. 

In this paper, we learn rich deep representations that are aligned across the three major natural modalities: vision, sound, and language.  We present a deep convolutional network that accepts as input either a sound, a sentence, or an image, and produces a representation shared across modalities. We capitalize on large amounts of in-the-wild data to learn this aligned representation across modalities. 
We develop two approaches that learn high-level representations that can be linked across modalities. Firstly, we use an unsupervised method that leverages the natural synchronization between modalities to learn an alignment. Secondly, we design an approach to transfer discriminative visual models into other modalities. 

\begin{figure*}[t]
    \centering
    \includegraphics[width=\linewidth]{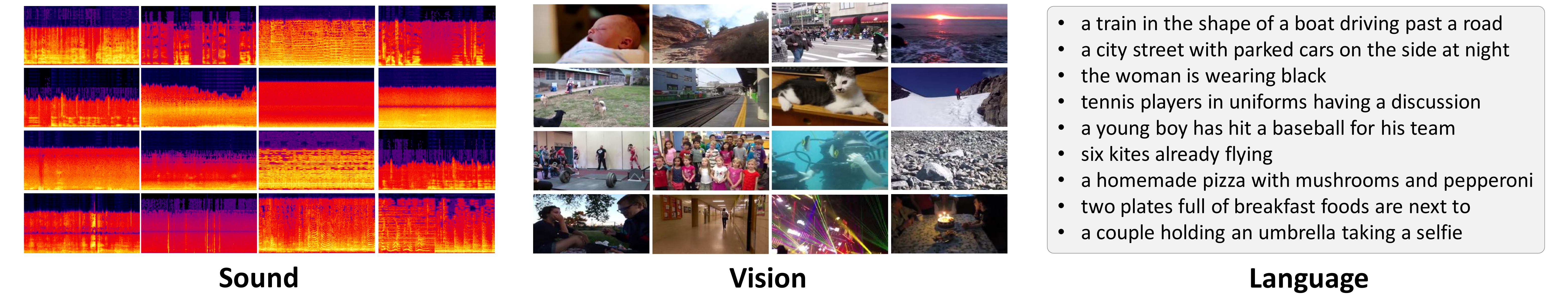}
    \caption{\textbf{Dataset:} We learn deep, aligned representations by capitalizing on large amounts of raw, unconstrained data.}
    \label{fig:dataset}
\end{figure*}

Our experiments and visualizations show that a representation automatically emerges that detects high-level concepts independent of the modality. Figure \ref{fig:teaser} visualizes this learned representation: notice how units  in the upper layers have learned automatically to detect some objects agnostic of the modality.
We experiment with this representation for several multi-modal tasks, such as cross-modal retrieval and classification. Moreover, although our network is only trained with image+text and image+sound pairs, our representation can transfer between text and sound as well, a transfer the network never saw during training.

Our primary contribution is showing how to leverage massive amounts of synchronized data to learn a deep, aligned cross-modal representation. While the methods in the paper are standard, their application on a large-scale to the three major natural modalities is novel to our knowledge. In the remainder of this paper, we describe the approach and experiments in detail. In section 2, we discuss our datasets and modalities. In section 3, we present a model for learning deep aligned cross-modal representations. In section 4, we present several experiments to analyze our representations. 

\subsection{Related Work}


\textbf{Vision and Sound:} Understanding the relationship between vision and sound has been recently explored in the computer vision community.
One of the early works, \cite{li2003multimedia}, explored the cross-modal relations between ``talking head'' images and speech through CCA and cross-modal factor analysis. \cite{zhang2007cross} applied CCA between visual and auditory features, and used common subspace features for aiding clustering in image-audio datasets. \cite{Song12} explored interaction between visual and audio modalities through human behavior analysis using Kernel-CCA and Multi-view Hidden CRF. \cite{ngiam2011multimodal} investigates RBM auto-encoders between vision and sound.
\cite{lampert2010weakly} investigated the relations between materials and their sound in a weakly-paired settings. Recent work  \cite{owens2015visually} has capitalized on material properties to learn to regress sound features from video, learn visual representations \cite{owens2016ambient}, and \cite{davis2014visual} analyzes small physical vibrations to recover sounds in video. We learn cross-modal relations from large quantities of unconstrained data. 


\textbf{Sound and Language:} Even though the relation between sound and language is mostly studied in the line of speech recognition \cite{rabiner1993fundamentals}, in this 
paper we are interested in matching sentences with auditory signals. This problem is mainly studied in the audio retrieval setting. Early work \cite{slaney2002semantic} performs semantic audio retrieval by aligning sound clusters with hierarchical text clusters through probabilistic models. \cite{chechik2008large} applies a passive-aggressive model for content-based audio retrieval from text queries.
\cite{turnbull2008semantic} uses probabilistic models for annotating novel audio tracks with words and retrieve relevant 
 tracks given a text-based query. However, we seek to learn the relationship between sound and language using vision as an intermediary, i.e. we do not use audio+text pairs.

\textbf{Language and Vision:} 
Learning to relate text and images has been extensively explored in the computer vision community. 
Pioneering work \cite{farhadi2010every,rashtchian2010collecting,ordonez2011im2text,kulkarni2013babytalk} explore image-captioning as a retrieval task. More recently,  \cite{vinyals2015show,karpathy2015deep,fang2015captions} developed deep large-scale models to generate captions from images. In this paper, rather than generating sentences, we instead seek to learn a representation that is aligned with images, audio, and text. \cite{farhadi2010every} explores aligned representations, but does not learn the representation with a deep architecture. Moreover, rather than using recurrent networks \cite{vinyals2015show}, we use convolutional networks for text. \cite{zhu2015aligning} learns to align books and moviesl. \cite{gong2014multi,gong2014improving} learn joint image-tag embeddings through several CCA variations. We instead seek to align three natural modalities using readily-available large-scale data. While \cite{gong2014multi} harnesses clusters of tags as a third view of the data, we instead obtain clusters from images through state-of-the-art visual categorization models. This is crucial since only the image modality is shared in both image+sound and image+text pairs.


\section{Datasets and Modalities}

We chose to learn aligned representations for sound, vision, and language because they are frequently used in everyday situations.  Figure \ref{fig:dataset} shows a few examples of the data we use.

\textbf{Sound:}
We are interested in natural environmental sounds. We download videos from videos on Flickr \cite{thomee2015yfcc100m} and extract their sounds. We downloaded over $750,000$ videos from Flickr, which provides over a year (377 days) of continuous audio, as well as their corresponding video frames.  The only pre-processing we do on the sound is to extract the spectrogram from the video files and subtract the mean. We extract spectrograms for approximately five seconds of audio, and keep track of the video frames for both training and evaluation. We use 85\% of the sound files for training, and the rest for evaluation.

\textbf{Language:} We combine two of the largest image description datasets available: COCO \cite{lin2014microsoft},
which contains $400,000$ sentences and $80,000$ images, and Visual Genome \cite{krishnavisualgenome}, which contains $4,200,000$ descriptions and $100,000$ images. The concatenation of these datasets results in a very large set of images and their natural language descriptions, which cover various real-world concepts. We pre-process the sentences by removing English stop words, and embedding each word with word2vec \cite{mikolov2013distributed}. 

\textbf{Images:} We use the frames from our sound dataset \cite{thomee2015yfcc100m} and the images from our language datasets \cite{lin2014microsoft,krishnavisualgenome}. In total, we have nearly a million images which are synchronized with either sound or text (but not both). The only pre-processing we do on the images is subtracting the channel-wise mean RGB value. We use the same train/test splits as their paired sounds/descriptions.

\textbf{Synchronization:}
We use the synchronous nature of these modalities to learn the relationships between them. We have pairs of images and sound (from videos) and pairs of images and text (from caption datasets). Note we lack pairs of sound and text during training. Instead, we hope our network will learn to map between sound and text by using images as a bridge (which our experiments suggest happens). To evaluate this, we also collected $1,000$ text descriptions of videos (image/sound) from workers on Amazon Mechanical Turk \cite{sorokin2008utility}, which we only use for testing the ability to transfer between sound and text.


\section{Cross-Modal Networks}

\begin{figure*}[t]
\centering
\includegraphics[width=\linewidth]{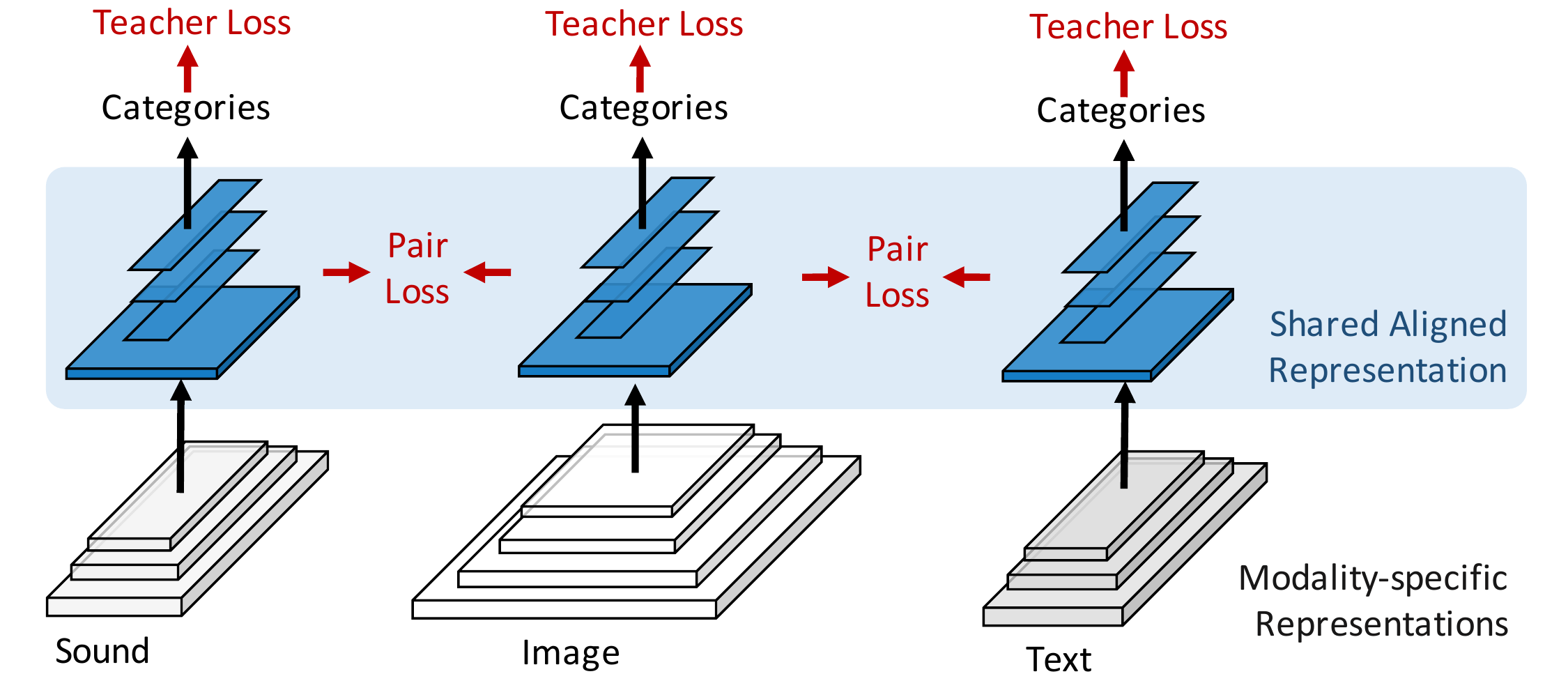}
\caption{\textbf{Learning Aligned Representations:} We design a network that accepts as input either an image, a sound, or a text. The model produces a common shared representation that is aligned across modality (blue) from modality-specific representations (grays). We train this model using both a model transfer loss, and a ranking pair loss. The modality-specific layers are convolutional, and the shared layers are fully connected.}
\label{fig:method}
\end{figure*}

We design a model that can accept as input either an image, a sound, or a sentence, and produces a common representation shared across modalities.  Let $x_i$ be a sample from modality $x$, and $y_i$ be the corresponding sample from modality $y$. For example, $x_i$ may be an image, and $y_i$ may be a sound of that image. For clarity, here we describe how to align two modalities $x$ and $y$, but the method easily generalizes to any number of modalities (we do three). 

Our goal is to learn representations for $x_i$ and $y_i$ that are aligned. We write $f_x(x_i)$ to be the representation in modality $x$, and $f_y(y_i)$ to be the representation in modality $y$.  Representations $f_x(x_i)$ and $f_y(y_i)$ are aligned if they are close to each other under some distance metric, e.g.\ cosine similarity. However, similarity alone is not enough because there is a trivial solution to ignore the input and produce a constant. Instead, we desire the representation to be both aligned and discriminative.  We explore two approaches.

\subsection{Alignment by Model Transfer}

We take advantage of discriminative visual models to teach a student model to have an aligned representation. Let $g(x_i)$ be a teacher model that estimates class probabilities for a particular modality. For example, $g(x_i)$ could be any image classification model, such as AlexNet \cite{simonyan2014very}.  Since the modalities are synchronized, we can train $f_y(y_i)$ to predict the class probabilities from the teacher model $g(x_i)$ in another modality. We use the KL-divergence as a loss:
\begin{align}
\sum_i^N D_{\textrm{KL}}\left(g(x_i) || f_y(y_i)\right)
\label{eqn:kl}
\end{align}
where $D_{\textrm{KL}(P || Q)} = \sum_j P_j \log \frac{P_j}{Q_j}$. This objective by itself will enable alignment to emerge at the level of categories predicted by $g$. However, the internal representations of $f$ would not be aligned since each student model is disjoint.

To enable an alignment to emerge in the internal representation, we therefore constrain the upper layers of the network to have shared parameters across modalities, visualized in Figure \ref{fig:method}. While the early layers of $f$ are specific to modality, the upper layers will now be shared. This encourages an internal representation to emerge that is shared across modalities. Interestingly, as we show in the experiments, visualizations suggest that hidden units emerge internally to detect some objects independent of modality.

Student-teacher models have been explored in transfer learning before \cite{aytar2016soundnet,gupta2015cross}. In this work, we are instead transferring into an aligned representation, which is possible by constraining the  learned parameters to be shared across the upper levels of representation. 

\subsection{Alignment by Ranking} 

We additionally employ a ranking loss function to obtain both aligned and discriminative representations:
\begin{align}
    \sum_i^N \sum_{j \ne i} \max\{0, \Delta - \psi(x_i, y_i) + \psi(x_i, y_j)\}
    \label{eqn:ranking}
\end{align}
where $\Delta$ is a margin hyper-parameter, $\psi$ is a similarity function, and $j$ iterates over negative examples. Note that, for clarity and in slight abuse of notation, $f$ may be a different layer in the network from above.

This loss seeks to push paired examples close together in representation space, and mismatched pairs further apart, up to some margin $\Delta$. We use cosine similarity in representation space:
\begin{align}
    \psi(x,y) = \textrm{cos}(f_x(x), f_y(y)) 
\end{align}
where $\textrm{cos}$ is the cosine of the angle between the two representation vectors. 

Ranking loss functions are commonly used in vision to learn cross-modal embeddings in images and text \cite{kiros2014unifying,socher2014grounded,karpathy2015deep,vendrov2015order,frome2013devise}. Here, we are leveraging them to learn aligned, discriminative representations across three major natural modalities using in-the-wild data.

\subsection{Learning}

To train the network, we use the model transfer loss in Equation \ref{eqn:kl} and the ranking loss in Equation \ref{eqn:ranking} on different layers in the network. For example, we can put the model transfer loss on the output layer of the network, and the ranking loss on all shared layers in the network. The final objective becomes a sum of these losses. 

\textbf{Model transfer:} We train student models for sound, vision, and text to predict class probabilities from a teacher ImageNet model. We constrain the upper weights to be shared in the student models. Since vision is a rich modality with strong recognition models, it is an attractive resource for transfer.

\textbf{Ranking:} We apply the ranking loss for alignment between vision $\rightarrow$ text, text $\rightarrow$ vision, vision $\rightarrow$ sound, and sound $\rightarrow$ vision on the last three hidden activations of the network. Since we do not have large amounts of sound/text pairs, we do not supervise those pairs. Instead, we expect the model to learn a strong enough alignment using vision as a bridge to enable transfer between sound/text (which our experiments suggest). 

\subsection{Network Architecture} 


Our network has three different inputs, depending on the modality of the data. We design each input to have its own disjoint pathway in the beginning in the network. In the end, however, the pathways converge to common layers that are shared across all modalities. Our intention is that the disjoint pathways can adapt to modal-specific features (such as shapes, audible notes, or text phrases), while the shared layers can adapt to modal-robust features (such as objects and scenes). 

\textbf{Sound Network:} The input to our sound pathway are spectrograms. Since sound is a one-dimensional signal, we use a four-layer one-dimensional convolutional network to transform the spectrogram into a higher-level representation. The output of the sound network is then fed into the modal-agnostic layers.
 
\begin{figure*}[tb]
    \centering
    \includegraphics[width=\linewidth]{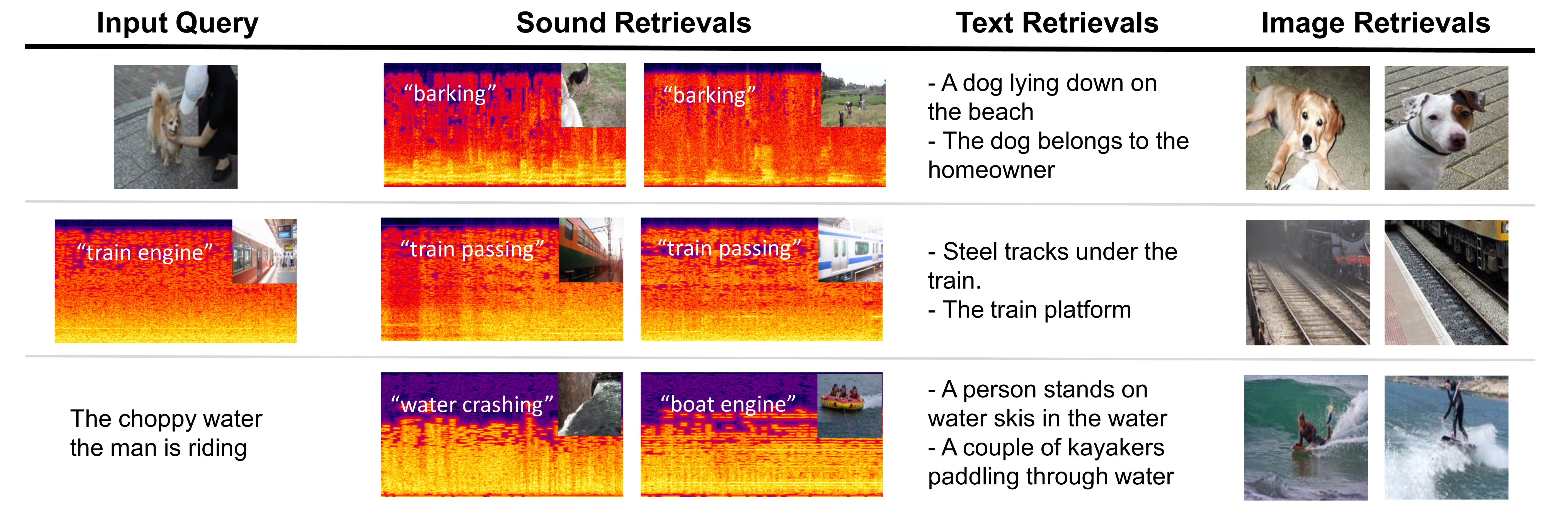}
    \vspace{-1em}
    \caption{\textbf{Example Cross-Modal Retrievals:} We show top retrievals for cross-modal retrieval between sounds, images, and text using our deep representation.}
    \label{fig:cmr-qual}
    \vspace{-1em}
\end{figure*}

\textbf{Text Network:}
The input to our text pathway are sentences where each word is embedded into a word representation using word2vec \cite{mikolov2013distributed}. By concatenating each word together, we can use a deep one-dimensional convolutional network on the sentence, similar to \cite{kim2014convolutional}. We again use a four-layer network. While the earlier layers in the network have a small receptive field and can only detect simple $n$-grams, by stacking convolutions, the later layers have a larger receptive field.
In contrast to \cite{kim2014convolutional} which uses convolutions of varying kernel size to handle long-range dependencies, we instead use convolutions with fixed kernel sizes, but go deeper to capture long-range dependencies between words. The output of this network is finally fed into the modal-agnostic layers.

\textbf{Vision Network:} 
Our visual network follows the standard Krizhevsky architecture \cite{krizhevsky2012imagenet}. We use the same architecture up until \texttt{pool5}, which is flattened and directly fed into the modal-agnostic layers. 

\textbf{Shared Network:}
The outputs from the sound, text, and vision networks are fixed length vectors with the same dimensionality. In order to create a representation that is independent of the modality, we then feed this fixed length vector into a network that is shared across all modalities, similar to \cite{lluis}. We visualize this sharing in Figure \ref{fig:method}. While the weights in the earlier layers are specific to their modality, the weights in the upper layers are shared across all modalities. We use two fully connected layers of dimensionality $4096$ with rectified linear activations as this shared network. The output is $1000$ dimensional with a softmax activation function.

\bgroup
\setlength{\tabcolsep}{7pt}
\begin{table}[t]
    \small
    \centering
    \begin{tabular}{l | c | c | c | c }
                & IMG & SND & IMG & TXT \\
                & $\downarrow$ & $\downarrow$ & $\downarrow$ & $\downarrow$ \\
         Method  & SND & IMG & TXT & IMG \\
        \hline
        Random & 500.0 & 500.0 & 500.0 & 500.00 \\
        Linear Reg. & 345.8 & 319.8 & 14.2 & 18.0  \\
        CCA \cite{rasiwasia2010new} & 313.6 & 316.1 & 17.0 & 16.2 \\
        Normalized CCA \cite{gong2014multi} & 295.6 & 296.0 & 14.2 & 12.8 \\
        \hline
        Ours: Model Transfer & 144.6 & 143.8 & 8.5 & 10.8 \\
        Ours: Ranking & 49.0 & {\bf 47.8} & 8.6 & 8.2 \\
        Ours: Both & {\bf 47.5} & 49.5 & {\bf 5.8} & {\bf 6.0} \\
    \end{tabular}

    \caption{\textbf{Cross Modal Retrieval:} We evaluate average median rank for cross-modal retrieval on our held-out validation set. Lower is better. See Section \ref{sec:cmr} for details.}
    \vspace{-1em}
    \label{tab:cmr}
\end{table}
\egroup

\begin{table}
        \centering
        \small
    \begin{tabular}{l | c | c }
         Method & TXT $\rightarrow$ SND & SND $\rightarrow$ TXT \\
        \hline
        Random & 500.0 & 500.0 \\
        Linear Reg. & 315.0 & 309.0\\
        \hline
        Ours: Model Transfer & 140.5 & 142.0 \\
        Ours: Ranking & 190.0 & 189.5 \\
        Ours: Both & {\bf 135.0} & {\bf 140.5} \\
    \end{tabular}
    \vspace{-1em}
    \caption{\textbf{Cross Modal Retrieval for Sound and Text:} We evaluate average median rank for retrievals between sound and text. See Section \ref{sec:sound2text}}
    \label{tab:cmr-soundtxt}
    \vspace{-1em}
\end{table}

\begin{table*}
\centering
\begin{tabular}{l|c c c|c c c|c c c}
Train Modality: & \multicolumn{3}{c|}{IMG} & \multicolumn{3}{c|}{SND} & \multicolumn{3}{c}{TXT} \\
Test Modality: & IMG & SND & TXT &  IMG & SND & TXT & IMG & SND & TXT \\
\hline
Chance & 2.3 & 2.3 & 2.3  & 2.3 & 2.3 & 2.3  & 2.3 & 2.3 & 2.3 \\
Linear Reg.  & 26.5 & 3.3 & 23.1 & 3.0 & 6.6 & 2.9 & 18.3 & 3.4 & 34.3\\
CCA TXT$\leftrightarrow$IMG & 23.8 & - & 22.2 & - & - & - & 18.5 & - & 35.6 \\
CCA SND$\leftrightarrow$IMG & 21.1 & 3.0 & - & 2.7 & 6.8 & - & - & - & - \\
\hline
Ours: Ranking & 23.5 & 5.7 & 21.3 & 6.6 & 5.7 & 6.3 & 11.3 & 5.2 & 32.9\\ 
Ours: Model Transfer  & 30.9 & 5.6 & 32.0 & 8.7 & {\bf 9.0} & 12.3 & 26.5 & 5.1 & 39.0\\ 
Ours: Both  & {\bf 32.6} & {\bf 5.8} & {\bf 33.8} & {\bf 12.8} & {\bf 9.0} & {\bf 15.2} & {\bf 22.6} & {\bf 6.2} & {\bf 40.3}
\end{tabular}
\caption{\textbf{Classifier Transfer:} We experiment with training scene classifiers in one modality, but testing on a different modality. Since our representation is aligned, we can transfer the classifiers without any labeled examples in the target modality. The table reports classification accuracy and the dash indicates a comparison is not possible because CCA only works with two views. The results suggest that our representation obtains a better alignment than baseline methods. Moreover, this shows that the representation is both aligned and discriminative.} 
\label{fig:cls-trans}
\end{table*}

\subsection{Implementation Details}

\textbf{Optimization:} We optimize the network using mini-batch stochastic gradient descent and back propagation \cite{lecun1998gradient}. We use the Adam solver \cite{kingma2014adam} with a learning rate of $0.0001$. We initialize all parameters with Gaussian white noise. We train with a batch size of $200$ for a fixed number of iterations ($50,000$). We train the network in Caffe \cite{jia2014caffe} and implement a new layer to perform the cosine similarity. Training typically takes a day on a GPU.

\textbf{Sound Details:} The input spectrogram is a $500 \times 257$ signal, which can be interpreted as $257$ channels over $500$ time steps. We use three one-dimensional convolutions with kernel sizes 11, 5, and 3 and 128, 256, 256 filters respectively. Between each convolutional layer, we use rectified linear units, and downsample with one-dimensional max-pooling by a factor of $5$. The output of these convolutions is a $4 \times 256$ feature map. Since these convolutions are over time and the other modalities do not have time (e.g., images are spatial), we finally project this feature map to a $9216$ dimensional vector with a fully connected layer, which is fed into the modality-agnostic layers. 

\textbf{Text Details:} The pretrained model for word2vec embeds each word into a $300$ dimensional vector. We concatenate words in a sentence into a fixed length matrix of size $16 \times 300$ for $16$ words. We pad shorter sentences with zeros, and crop longer sentences, which we found to be effective in practice. We then have three one-dimensional convolutions with $300$ filters and kernel size of $3$ with rectified linear activation functions. We have max-pooling after the second and third convolutions to down-sample the input by a factor of two. We finally have a fully connected layer to produce a $9216$ dimensional vector that is fed into the shared layers.


\begin{figure*}[t!]
    \centering
    \includegraphics[width=\linewidth]{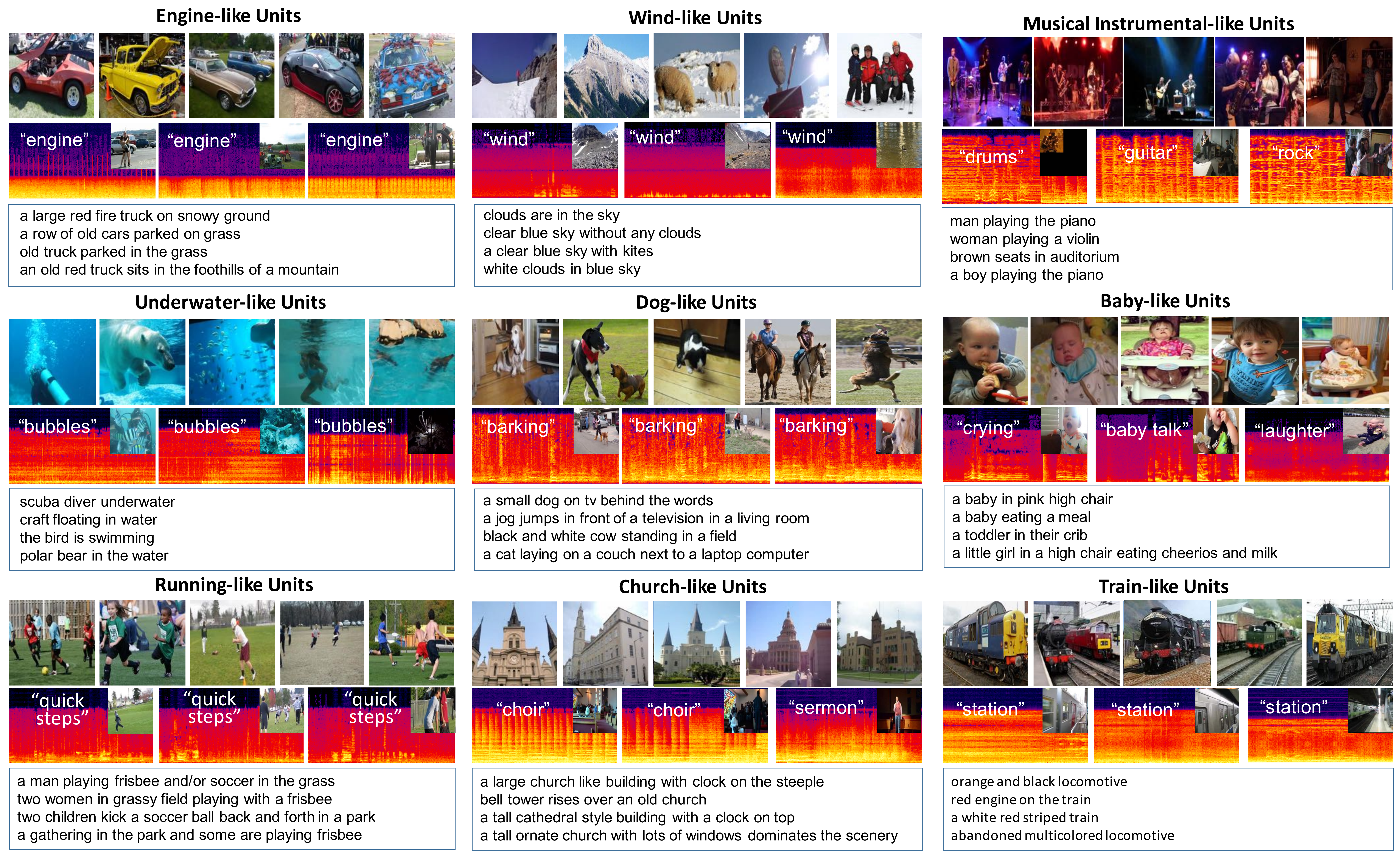}
    \vspace{-2em}
    \caption{\textbf{Hidden Unit Visualization:} We visualize a few units from the last hidden layer of our cross modal network. Note that, on top of the spectrograms (yellow/red heatmaps), we also show the original video and a blurb to describe the sound, which is only included for visualization purposes. See Section \ref{sec:vis}.}
    \label{fig:neuron}
\end{figure*}

\section{Experiments}

We present three main experiments to analyze our aligned representation. Firstly, we experiment with cross-modal retrieval that, given a query in one modality, find similar examples in all modalities. Secondly, we show discriminative classifiers trained on our representation transfer to novel modalities. Finally, we visualize the internal representation, and show that some object detectors independent of modality are automatically emerging. 

\subsection{Experimental Setup}

We split our data into disjoint training, validation, and testing sets. We learn all models on training, fit hyper-parameters on the validation set, and report performance on the testing set. For both validation and testing, we use $5,000$ video/sound pairs and similarly $5,000$ image/text pairs. The rest is used for training. 
For the image descriptions, we use the standard training/validation split from COCO \cite{lin2014microsoft}. Since Visual Genome \cite{krishnavisualgenome} did not release a standard training/validation split, we randomly split the dataset. However, because Visual Genome has some overlap with the images in COCO, if an image belongs in both COCO and Visual Genome, then we assigned it to the same training/validation/testing split as COCO in order to keep the splits disjoint. We train all networks from scratch (random initialization).

\subsection{Cross Modal Retrieval}
\label{sec:cmr}

We quantify the learned alignment by evaluating our representations at a cross-modal retrieval task. Given a query input in one modality, how well can our representation retrieve its corresponding pair from a different modality? For our method, we input the example from the query modality into our network, and extract the features from the last hidden layer. We then normalize the query features to be zero mean and unit variance. Finally, we find examples in the target modality with the nearest cosine similarity.

We compare against two baselines for this task. 

\textbf{CCA:} Firstly, we compare against CCA \cite{rasiwasia2010new}, which is a state-of-the-art method for cross-modal retrieval. Using our training set, we compute CCA between images and text, and CCA between images and sound. To do this, we need to operate over a feature space. For images, we use \texttt{fc7} features from \cite{krizhevsky2012imagenet}. For sentences, we use a concatenation of words embedded with word2vec \cite{mikolov2013distributed}. For sound, we reduce the dimensionality of the spectrograms to $512$ dimensions using PCA, which we found improved performance. We do retrieval using the joint latent space learned by CCA.  

\textbf{Linear Regression:} Secondly, we compare against a linear regression trained from the query modality to visual features, and use vision as the common feature space. We use the same features for linear regression as we did in the CCA baseline. Note we add a small isotropic prior to the transformation matrix which acts as a regularizer. We then perform retrieval using the regressed target features using cosine similarity.


\textbf{Results:} Using our test set, we report the average median rank over five splits of $1,000$ each, following \cite{vendrov2015order}. Table \ref{tab:cmr} shows that our representation learns a significantly better alignment across vision, sound, and text than baselines. However, for retrieval between text and images, our method marginally outperforms CCA. Since our network is capable of learning deep features, our method can learn to align spectrograms using features higher level than what is possible with CCA. On the other hand, since text is already high-level, our method only provides a slight advantage over CCA for text/images. In general, the task of retrieving between images and sound appears to be a more challenging task than between images and text, perhaps because less information is available in sound and the original features are not as high level. We show some qualitative top retrievals in Figure \ref{fig:cmr-qual}.

\subsection{Sound and Text Transfer}
\label{sec:sound2text}
Although our network was trained using only image/sound and image/text pairs, we also experiment with
transfer between sound/text. The network never saw sound/text pairs during training. This task is particularly challenging because the network would need to develop a strong enough alignment between modalities such that it can exploit images as a bridge between sound and text. 

\textbf{Baseline:} Although we cannot train a linear regression between sound and text (because there are no pairs), we can train linear regressions from spectrograms to image features, and text features and image features. We can use the regressed image features as the common space to do retrieval.

For our method, we simply perform retrieval using cosine similarity using our learned representations. Given a sound query, we compute its representation, and retrieve text that is near it, and similarly for the reverse direction.

\textbf{Results:} Table \ref{tab:cmr-soundtxt} reports the average median rank for sound/text retrievals. Our experiments suggest that deep cross-modal representations outperform both cluster CCA and a linear regression by considerable margins (over $100$ points). We believe this is the case because our network is capable of learning high-level features, which are easier to align across modalities. Interestingly, our network can transfer between sound/text only slightly worse than sound/images, suggesting that our network is capable of learning alignment between modalities even in the absence of synchronized data. 


\subsection{Zero Shot Classifier Transfer}

We explore using the aligned representation as a means to transfer classifiers across modalities. If the representation obtains a strong enough alignment, then an object recognition classifier trained in a source modality should still be able to recognize objects in a different target modality, even though the classifier never saw labeled examples in the target modality.

\textbf{Dataset:} To quantify performance on this task, we collected a new medium size dataset for transferring classifiers across vision, sound, an text modalities. We annotate held-out videos into $42$ categories consisting of objects and scenes using Amazon Mechanical Turk. The training set is $2,799$ videos and the testing set is $1,050$ videos, which is balanced. We additionally annotated each video with a short text description, similar to sentences from COCO. This results in a dataset where we have paired data across all modalities. 

\textbf{Classifier:} We experiment with training a linear one-vs-all SVM to recognize the categories where we train and test on different modalities using our aligned representation as the feature space. Note to pick hyper-parameters, we use two-fold cross validation on the training set.

\textbf{Results:} Table \ref{fig:cls-trans} reports classification accuracy for the classifier across modalities. We compare the representation from our approach versus a representation obtained by CCA and Linear regression, similar to before. Our experiments suggest that our representation learns a stronger discriminative alignment than CCA and Linear regression, obtaining up to 10\% gain over baselines.

Particularly the cross-modal columns  in table \ref{fig:cls-trans}, where train and test modalities are different, shows that even without seeing any example from the target modality our methods can achieve significant classification accuracies. The most challenging source modality for training is sound, which makes sense as vision and text are very rich modalities. However, our approach still learns to align sound with vision and text. By combining both a paired ranking objective and a model transfer objective, our representation is both discriminative and aligned.


\subsection{Visualization}
\label{sec:vis}

To better understand what our model has learned, we visualize the hidden units in the shared layers of our network, similar to \cite{zhou2014object}. Using our validation set, we find which inputs activate a unit in the last hidden layer the most, for each modality. We visualize the highest scoring inputs for several hidden units in Figure \ref{fig:neuron}. We observe two properties. Firstly, although we do not supervise semantics on the hidden layers, many units automatically emerge that detect high-level concepts. Secondly, many of these units seem to detect objects independently of the modality, suggesting the representation is learning an alignment at the object level.

\section{Conclusion}

Invariant representations enable computer vision systems to operate in unconstrained, real-world environments. We believe aligned, modality-robust representations are crucial for the next generation of machine perception as the field begins to leverage cross-modal data, such as sound, vision, and language. In this work, we present a deep convolutional network for learning cross-modal representations from over a year of video and millions of sentences. Our experiments show an alignment emerges that improves both retrieval and classification performance for challenging in-the-wild situations. Although the network never saw pairs of sounds and text during training, our experiments empirically suggest it has learned an alignment between them, possibly by using images as a bridge internally. Our visualizations reveal that units for high-level concepts emerge in our representation, independent of the modality. 



{
\small
\bibliographystyle{ieee}
\bibliography{egbib}
}

\end{document}